\documentclass[conference]{IEEEtran}
\IEEEoverridecommandlockouts
\usepackage{cite}
\usepackage{amsmath,amssymb,amsfonts}
\usepackage{enumitem}
\usepackage{algorithmic}
\usepackage{graphicx}
\usepackage{textcomp} 
\usepackage{soul}
\usepackage[dvipsnames]{xcolor}
\definecolor{pathwaycolor}{RGB}{255, 230, 204}
\definecolor{gocolor}{RGB}{204, 255, 229}
\definecolor{structurecolor}{RGB}{204, 229, 255}
\newcommand{\hlpathway}[1]{\sethlcolor{pathwaycolor}\hl{#1}}
\newcommand{\hlgo}[1]{\sethlcolor{gocolor}\hl{#1}}
\newcommand{\hlstruct}[1]{\sethlcolor{structurecolor}\hl{#1}}
\usepackage{booktabs}
\usepackage[normalem]{ulem}
\usepackage{multirow}
\usepackage{graphicx}

\def\BibTeX{{\rm B\kern-.05em{\sc i\kern-.025em b}\kern-.08em
    T\kern-.1667em\lower.7ex\hbox{E}\kern-.125emX}}
\begin{document}
\title{CoTox: Chain-of-Thought-Based Molecular Toxicity Reasoning and Prediction
\thanks{\noindent\rule{\linewidth}{0.4pt}\\
*Corresponding author.\\
}
}


\author{
\IEEEauthorblockN{
Jueon Park\textsuperscript{1},
Yein Park\textsuperscript{1},
Minju Song\textsuperscript{1},
Soyon Park\textsuperscript{1},
Donghyeon Lee\textsuperscript{1,2},
Seungheun Baek\textsuperscript{1},\\
Jaewoo Kang\textsuperscript{1,2,$\ast$}
}
\IEEEauthorblockA{
\textsuperscript{1}Department of Computer Science and Engineering, Korea University, Seoul 17035, Republic of Korea\\
\textsuperscript{2}AIGEN Sciences, Seoul 04778, Republic of Korea\\
\{jueon\_park, 522yein, minjusong, soyon\_park, dong9733, sheunbaek, kangj\}@korea.ac.kr
}
}

\maketitle

\begin{abstract}
Drug toxicity remains a major challenge in pharmaceutical development. Recent machine learning models have improved \textit{in silico} toxicity prediction, but their reliance on annotated data and lack of interpretability limit their applicability. This limits their ability to capture organ-specific toxicities driven by complex biological mechanisms. Large language models (LLMs) offer a promising alternative through step-by-step reasoning and integration of textual data, yet prior approaches lack biological context and transparent rationale. To address this issue, we propose CoTox, a novel framework that integrates LLM with chain-of-thought (CoT) reasoning for multi-toxicity prediction. CoTox combines chemical structure data, biological pathways, and Gene Ontology (GO) terms to generate interpretable toxicity predictions through step-by-step reasoning. Using GPT-4o, we show that CoTox outperforms both traditional machine learning and deep learning model. We further examine its performance across various LLMs to identify where CoTox is most effective. Additionally, we find that representing chemical structures with IUPAC names, which are easier for LLMs to understand than SMILES, enhances the model’s reasoning ability and improves predictive performance. To demonstrate its practical utility in drug development, we simulate the treatment of relevant cell types with drug and incorporated the resulting biological context into the CoTox framework. This approach allow CoTox to generate toxicity predictions aligned with physiological responses, as shown in case study. This result highlights the potential of LLM-based frameworks to improve interpretability and support early-stage drug safety assessment. 
The code and prompt used in this work are available at https://github.com/dmis-lab/CoTox.  
\\
\end{abstract}


\section{Introduction}

Toxicity has become a major cause of failure in drug development, frequently resulting in the termination of candidate compounds during development \cite{sun202290}. More critically, undetected toxicity can lead to severe adverse effects in patients after market approval, necessitating drug withdrawals \cite{babai2021safety}. Such failures incur substantial financial losses, waste valuable resources, and undermine the pharmaceutical industry's reputation. Consequently, early and accurate toxicity prediction has become a critical priority in drug development.

Recent advances in artificial intelligence (AI) have extended its application to drug toxicity prediction. Machine learning (ML) and deep learning (DL) models have shown promise in identifying toxic compounds based on molecular features\cite{tran2023artificial}. However, these models typically rely on large amounts of experimentally annotated toxicity data for training and often lack interpretability, limiting their utility in mechanistic understanding and decision-making.

To overcome these limitations, large language models (LLMs) have emerged as a promising alternative, offering the ability to perform contextual reasoning and integrate diverse types of information. In particular, the introduction of advanced prompting and reasoning techniques has brought their problem-solving ability closer to human-like thinking \cite{patil2025advancing}. This progress has led to growing interest in applying LLMs to drug discovery tasks \cite{zheng2024large}. Recent studies have begun exploring the use of general-purpose LLMs for toxicity prediction, demonstrating that these models can infer specific toxic effects, such as cardiotoxicity or osteotoxicity, directly from molecular representations using prompt-based approaches\cite{yang2025large, chen2025application}. These early efforts highlight the potential of LLMs in \textit{in silico} toxicology and set the stage for broader applications in drug safety evaluation.

Despite their promise, existing LLM-based toxicity prediction studies face several important limitations. First, they typically use SMILES strings, which are text-based representations of molecular structures, as input, expecting the model to infer toxicity directly from structural features. However, general-purpose LLMs are primarily trained on natural language and often struggle to fully understand the syntax and semantics of SMILES, which can limit their ability to capture subtle structural features relevant to toxicity \cite{jang2025improving}. Second, these studies rely solely on structural information while neglecting biological context, such as the pathways through which a drug interacts with the body. This is critical because organ-specific toxicity can result not only from on-target effects but also from off-target interactions with unintended biological pathways\cite{lin2019off}. Therefore, incorporating such context is essential for accurate prediction. Third, although LLMs are capable of generating predictions through explicit reasoning, existing approaches do not leverage this ability. They offer no explanation of how predictions are derived, which limits interpretability and reduces trust in applications such as drug development.

To address these limitations, we propose CoTox, a framework that enhances toxicity prediction by integrating both structural and biological information into the input of LLM. Unlike previous approaches that use only chemical structure features, CoTox incorporates additional biological process, including pathway involvement and Gene Ontology (GO) terms, which are known to play a crucial role in mediating drug-induced toxicity \cite{chen2016analysis}. Furthermore, we adopt the Chain-of-Thought (CoT) prompting strategy\cite{wei2022chain}, which enables the model to perform step-by-step reasoning by sequentially processing structural and biological information. This approach allows the model to generate predictions based on a more comprehensive and interpretable understanding of the underlying toxicity mechanisms. In addition, CoTox uses molecular structure not in the form of SMILES strings, which are difficult for general-purpose LLMs to interpret, but rather as IUPAC names, the standardized nomenclature used throughout the scientific community. Since IUPAC names are more human-readable than other chemical formats, they can serve as a bridge between formal encodings and natural language, helping the model better capture molecular structures\cite{ramos2025review}. By adopting IUPAC names, CoTox enables LLMs to better interpret molecular structures for toxicity prediction, with performance gains observed in most models, especially those using reasoning-based approaches.

In summary, CoTox enhances toxicity prediction by integrating chemical structure in IUPAC format and biologically relevant features such as pathways and GO terms, combined with stepwise reasoning for improved interpretability.

\section{Related Work}
Large language models (LLMs) have recently been applied to molecular toxicity prediction by leveraging their capacity for general reasoning and language understanding. For example,~\cite{yang2025large} utilized ChatGPT~\cite{chatgpt} to infer cardiotoxicity directly from SMILES strings through prompt-based binary classification. Similarly,~\cite{chen2025application} investigated osteotoxicity using DeepSeek and ChatGPT, prompting the models to explain structural causes of toxicity. These studies demonstrate that LLMs can generalize from molecular representations to predict specific toxicity types without fine-tuning.

Common to these studies is the use of purely structural inputs and the framing of toxicity prediction as a zero- or few-shot classification task. While this approach is attractive for its simplicity, it often lacks biological grounding. Moreover, the molecular inputs are not optimized for LLMs, which are not trained to interpret chemical syntax like SMILES\cite{jang2025improving}. As a result, prior approaches primarily rely on implicit pattern recognition, offering limited interpretability and failing to incorporate structured reasoning processes.

In contrast, recent advances in related scientific domains such as chemistry and biology have begun to incorporate structured reasoning into LLM frameworks. For instance, \textsc{S\lowercase{truct}C\lowercase{hem}}\cite{ouyang2023structured} proposed a multi-phase prompting strategy to solve complex chemistry problems through formula extraction, compositional reasoning, and confidence-based refinement. MolRAG\cite{xian2025molrag} enhanced molecular property prediction by combining retrieval-augmented generation with chain-of-thought prompting, grounding predictions in structurally similar molecules. In the biological domain, \textsc{B\lowercase{io}R\lowercase{eason}}\cite{fallahpour2025bioreason} integrated a DNA foundation model with an LLM to enable multi-step reasoning over genomic sequences, producing interpretable deductions in tasks such as pathway prediction and variant effect analysis.

Inspired by these developments, our work introduces both chemical and biological context along with stepwise reasoning into the LLM-based toxicity prediction pipeline. Whereas previous approaches relied solely on structural inputs such as SMILES, our framework incorporates biological process information to guide toxicity prediction, and adopts IUPAC names to express molecular structures in a more interpretable and human-readable format. By integrating structured prompts and interpretable biological cues, CoTox moves beyond surface-level classification, enabling mechanistic and biologically grounded inference in LLM-based toxicology.

\section{Method}
\begin{figure*}[!t]
    \centering
    \vspace{5pt}
    \includegraphics[width=0.9\textwidth]{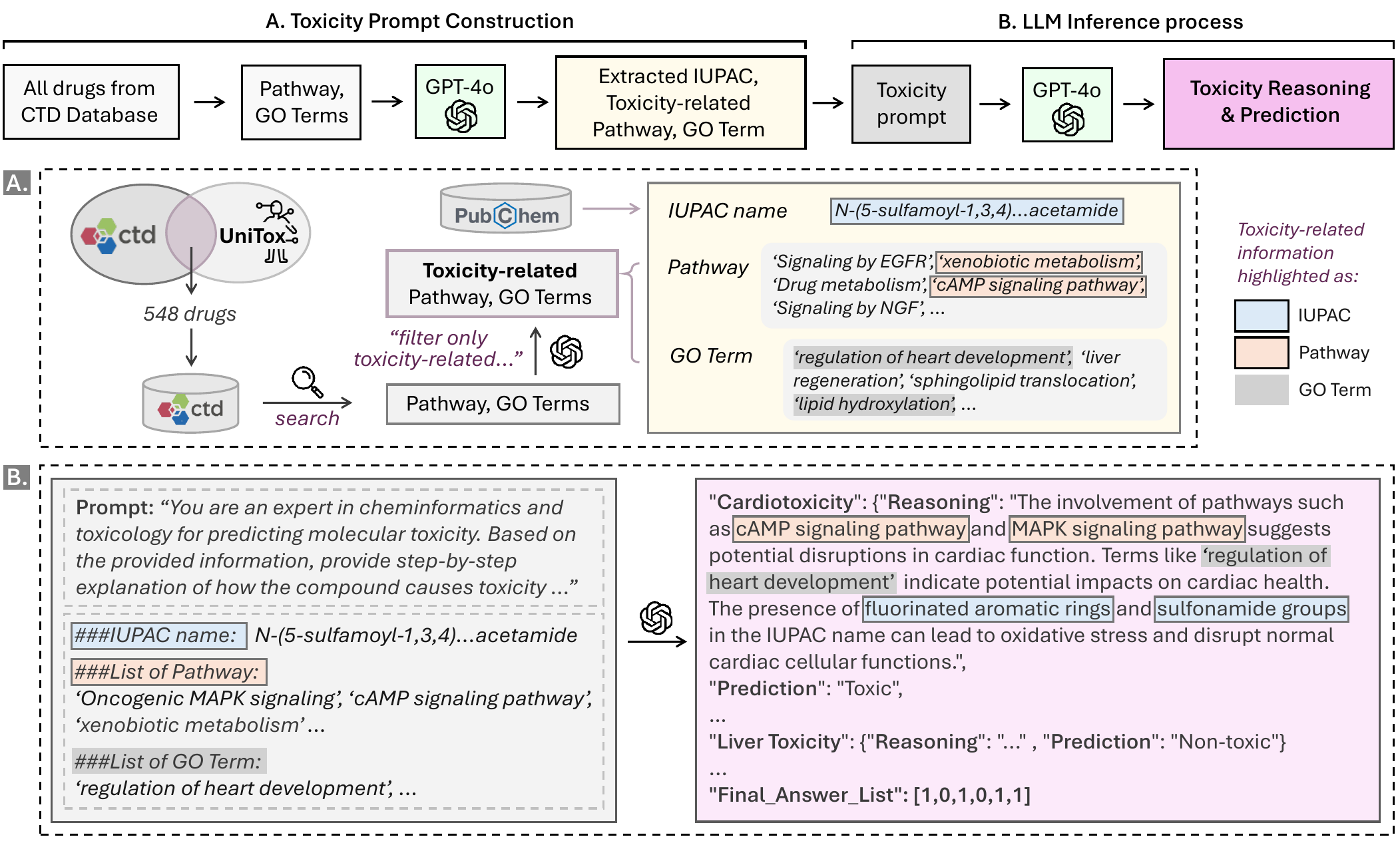}
    \caption{\textbf{Overview of CoTox Framework}, (A) Toxicity Prompt Construction, where toxicity-related pathways and GO terms are retrieved from the CTD and additional extracting toxicity-related information is carried out by GPT-4o, followed by integration with IUPAC for toxicity prediction.
    (B) LLM inference process using a structured toxicity prompt to generate reasoning and predictions based on extracted biological data and chemical structure.}
    \vspace{-5pt}
    \label{fig:main_fig}
\end{figure*}

\subsection{Toxicity Prompt Construction}
We designed structured prompts that incorporate both biological and chemical context for each compound to support LLM-based reasoning for toxicity prediction. This process involved retrieving relevant biological processes and obtaining standardized, human-readable chemical names that can serve as meaningful input for the language model.

To obtain the biological information, we extracted pathway and Gene Ontology (GO) annotations from the Comparative Toxicogenomics Database (CTD)\cite{davis2025comparative}, which provides curated associations between chemicals, genes, pathways, and GO terms. Since the full list of CTD annotations includes many biological processes unrelated to toxicity, we employed GPT-4o to semantically filter relevant entries. A system prompt instructed the model to retain only toxicity-related pathways and GO terms. This allowed us to isolate biologically meaningful features for each compound that could support reasoning about toxic effects.
For the chemical context, we retrieved the IUPAC name of each compound using the PubChemPy, a Python wrapper for the PubChem PUG REST API\cite{kim2018update}. Given a drug name, PubChemPy queries the PubChem database and returns the corresponding IUPAC name. 

At the end of this process, each prompt was constructed using the IUPAC name of the compound along with the filtered list of toxicity-related pathways and GO terms. This dual-context input served as the basis for LLM reasoning, enabling the model to make predictions informed by both structural and biological insights.

\subsection{LLM Inference with Toxicity Prompting}
To integrate both biological and chemical contexts into the LLMs, we developed a tailored prompt based on a fundamental chat template structure.

In the “System Prompt,” the LLM was instructed to assume the role of a cheminformatics and toxicology expert. It was guided to predict compound toxicity based on three sources of information: pathway associations with toxicity mechanisms, biological implications of GO terms, and structural interpretation from IUPAC names. To ensure consistency and prevent irrelevant outputs, the model was required to respond in a strict JSON format without any additional commentary.
\setlength{\parskip}{0pt}
\setlength{\parindent}{1em}

In the “User Prompt,” we guided the model to predict the presence or absence of six organ-specific toxicity types by producing binary outputs of either “Toxic” or “Non-toxic”. To encourage mechanistic and interpretable reasoning, the prompt instructed the model to follow a four-step analytical process for each toxicity type. First, it examined the input pathways to determine their relevance to toxicity mechanisms. Second, it analyzed the associated GO terms to interpret the biological processes and molecular functions affected. Third, it used the IUPAC name of the compound to identify structural features that may support or explain the biological associations. Finally, the model synthesized information from pathways, GO terms, and the chemical structure into a coherent explanation describing how the compound may induce toxicity in each organ system. All reasoning steps and final predictions were returned in a standardized JSON format, ensuring consistency and interpretability across all outputs. Fig.\ref{fig:main_fig}. illustrates the overall workflow of this process.
\section{Experiments}
\subsection{Benchmark Dataset}
We utilized the UniTox dataset \cite{silberg2024unitox} to evaluate multi-organ toxicity predictions, focusing on six toxicity types: cardiotoxicity, hematological toxicity, infertility, liver toxicity, pulmonary toxicity, and renal toxicity. Two toxicity types—dermatotoxicity and ototoxicity—were excluded due to severe class imbalance. UniTox was constructed by applying GPT-4o to 2,418 FDA drug labels, where the model was prompted to summarize toxicity-related evidence and assign toxicity ratings in both ternary (No/Less/Most) and binary (Yes/No) formats. The resulting labels were validated against established FDA datasets (e.g., DICTrank, DILIrank) and clinician review, ensuring the reliability and interpretability of the dataset. For our evaluation, we adopted the binary (Yes/No) toxicity labels.
\useunder{\uline}{\ul}{}

\begin{table*}[ht]
\caption{F1-Score Comparison of Toxicity Prediction Prompts Using GPT-4o and ML/DL Models Across Toxicity Types}

\label{tab:toxicity_comparison}
\centering
\small
\setlength{\tabcolsep}{5pt}
\begin{tabular}{llccccccc}
\specialrule{1.5pt}{0pt}{4pt}
\textbf{Context} & \textbf{Method} & \textbf{Cardio} & \textbf{Hemato} & \textbf{Infertility} & \textbf{Liver} & \textbf{Pulmonary} & \textbf{Renal} & \textbf{Average} \\
\specialrule{0.2pt}{4pt}{4pt}
\multirow{8}{*}{\shortstack[l]{Chemical \\Structure}}
& XGBoost (ML)         & 0.673 & 0.779 & 0.479 & 0.648 & 0.452 & 0.427 & 0.576 \\
& Chemprop (DL)        & 0.663 & 0.775 & \uline{0.542} & 0.721 & 0.447 & \textbf{0.566} & 0.619 \\
\cmidrule(lr){2-9}
& SMILES-Zeroshot      & 0.486 & 0.186 & 0.324 & \uline{0.769} & 0.030 & 0.429 & 0.370 \\
& SMILES-Fewshot & 0.498 & 0.397 & 0.274 & 0.769 & 0.151 & 0.515 & 0.434 \\
& SMILES-CoT           & 0.523 & 0.492 & 0.234 & 0.729 & 0.053 & 0.426 & 0.409 \\
& IUPAC-Zeroshot       & 0.495 & 0.226 & 0.371 & 0.687 & 0.025 & 0.406 & 0.368 \\
& IUPAC-Fewshot & 0.454 & 0.386 & 0.242 & 0.697 & 0.117 & 0.519 & 0.402 \\
& IUPAC-CoT            & 0.476 & 0.521 & 0.301 & 0.693 & 0.070 & 0.442 & 0.417 \\
\specialrule{0.2pt}{4pt}{6pt}
\vspace{-1mm}
\parbox[c][1.9em][t]{2.5cm}{Biological Process}
& \parbox[c][1.9em][t]{2.5cm}{BioProcess-CoT} 
& \parbox[c][1.9em][t]{0.7cm}{0.684} 
& \parbox[c][1.9em][t]{0.7cm}{0.792} 
& \parbox[c][1.9em][t]{0.7cm}{0.496} 
& \parbox[c][1.9em][t]{0.7cm}{0.764} 
& \parbox[c][1.9em][t]{0.7cm}{0.531} 
& \parbox[c][1.9em][t]{0.7cm}{0.475} 
& \parbox[c][1.9em][t]{0.7cm}{0.624} 
\vspace{-1mm} \\
\specialrule{0.2pt}{3pt}{4pt}
\multirow{2}{*}{\shortstack[c]{Chem + Bio}}
& CoTox (SMILES)        & \textbf{0.723} & \uline{0.809} & 0.530 & \textbf{0.774} & \textbf{0.554} & \uline{0.564} & \uline{0.659} \\
& CoTox         & \uline{0.711} & \textbf{0.817} & \textbf{0.582} & 0.768 & \uline{0.541} & 0.557 & \textbf{0.663} \\
\specialrule{1.5pt}{3pt}{0pt}
\multicolumn{9}{r}{\footnotesize\textit{\textbf{Bold} = highest, \uline{Underlined} = second-highest}}
\end{tabular}
\end{table*}

\subsection{Experimental settings}
A multi-label classification task was conducted to predict six types of toxicity for a single chemical compound using the UniTox dataset. We split the dataset into training and test set, where the test set consisted of 548 compounds for which biological context information was available through the Comparative Toxicogenomics Database (CTD). The remaining compounds were used to train the baseline models. As a baseline, we first evaluated traditional machine learning and deep learning models. XGBoost\cite{chen2016xgboost} is a gradient boosting decision tree method that is widely used for tabular data. It has been frequently applied in toxicity prediction tasks. Chemprop\cite{heid2023chemprop} is a deep learning model based on a graph-based directed message passing neural network (D-MPNN). It learns molecular representations from SMILES strings and has been widely used for molecular property prediction through representation learning. 

We then compared the model’s performance using four different prompting strategies applied to GPT-4o~\cite{gpt4o}. In the Zeroshot setting, the model received only the molecular input, either SMILES or IUPAC, and was asked to predict toxicity without any prior examples or reasoning. The Fewshot setting included four example input and output pairs alongside the target molecule to guide the model's prediction through in-context learning. In the Chain-of-Thought (CoT) setting, the model was prompted to provide step-by-step reasoning before making its prediction, encouraging structural interpretation. Lastly, the CoTox setting extended CoT by incorporating pathway and GO term annotations, enabling the model to reason over both chemical structure and biological processes for a more comprehensive toxicity assessment.

To evaluate prompt-based methods, we used the selected 548-compound test set. For each toxicity type, we computed the F1-score by averaging the results across the five splits. We chose F1-score as it is a commonly used metric for binary classification tasks, effectively balancing precision and recall. We reported both the F1-scores for each of the six toxicity types and the overall average, calculated as the mean of these six scores. All prompt-based evaluations were conducted using GPT-4o to isolate the effect of prompt design from model architecture.

\vspace{4pt}
\section{Results \& Discussion}
\begin{table*}[ht]
\caption{F1-Score Comparison of LLM Performance Using the CoTox Prompt Across Toxicity Types}
\label{tab:llm_toxicity}
\centering
\small  
\setlength{\tabcolsep}{5pt}
\begin{tabular}{lccccccccc}
\specialrule{1.5pt}{0pt}{4pt}
\textbf{\parbox[c][0.8em][t]{1cm}{Model}} & 
\textbf{\parbox[c][0.8em][t]{1cm}{Cardio}} & 
\textbf{\parbox[c][0.8em][t]{1.1cm}{Hemato}} & 
\textbf{\parbox[c][0.8em][t]{1.3cm}{Infertility}} & 
\textbf{\parbox[c][0.8em][t]{0.8cm}{Liver}} & 
\textbf{\parbox[c][0.8em][t]{1.5cm}{Pulmonary}} & 
\textbf{\parbox[c][0.8em][t]{0.8cm}{Renal}} & 
\textbf{\parbox[c][0.8em][t]{1.05cm}{Average}} & 
\textbf{\parbox[c][2.0em][t]{1.4cm}{\shortstack{Average\\ (SMILES)}}} & 
\textbf{\parbox[c][2.0em][t]{0.7cm}{\shortstack{Gap\\ (\%)}}} \\
\specialrule{0.4pt}{4pt}{4pt}
\multicolumn{10}{l}{\textbf{General LLM w{/} CoTox}} \\
GPT-4o~\cite{gpt4o}         & 0.711 & 0.817 & 0.582 & 0.768 & 0.541 & 0.557 & 0.663 & 0.659 & \textcolor{ForestGreen}{0.61} \\
Llama3.1-8B~\cite{Llama3.1}    & \uline{0.738} & 0.823 & \uline{0.595} & 0.774 & \uline{0.586} & \uline{0.591} & \uline{0.685} & \uline{0.666} & \textcolor{ForestGreen}{2.78} \\
Llama3.1-70B~\cite{Llama3.1}   & 0.735 & \textbf{0.835} & 0.292 & 0.769 & 0.391 & 0.527 & 0.591 & 0.615 & \textcolor{RubineRed}{-3.85} \\
\specialrule{0.4pt}{4pt}{4pt}
\multicolumn{10}{l}{\textbf{Expert LLM w{/} CoTox}} \\
TxGemma-9B-Chat~\cite{wang2025txgemma}         & 0.444 & 0.485 & 0.383 & 0.512 & 0.318 & 0.382 & 0.421 & 0.387 & \textcolor{ForestGreen}{8.66} \\
\specialrule{0.4pt}{4pt}{4pt}
\multicolumn{10}{l}{\textbf{Reasoning LLM w{/} CoTox}} \\
o3~\cite{o3}         & 0.721 & 0.749 & 0.578 & \uline{0.778} & 0.518 & 0.554 & 0.650 & 0.562 & \textcolor{ForestGreen}{15.60} \\
DeepSeek-R1~\cite{deepseek-r1}    & 0.558 & 0.639 & 0.269 & 0.739 & 0.405 & 0.446 & 0.509 & 0.462 & \textcolor{ForestGreen}{10.24} \\
Qwen3-32B~\cite{qwen3}      & 0.634 & 0.778 & 0.214 & 0.768 & 0.449 & 0.512 & 0.559 & 0.486 & \textcolor{ForestGreen}{15.02} \\
Gemini-2.5-Pro~\cite{gemini2.5}  & \textbf{0.746} & \uline{0.831} & \textbf{0.630} & \textbf{0.794} & \textbf{0.591} & \textbf{0.606} & \textbf{0.700} & \textbf{0.698} & \textcolor{ForestGreen}{0.21} \\
\specialrule{1.5pt}{3pt}{0pt}
\multicolumn{10}{r}{\footnotesize\textit{\textbf{Bold} = highest, \uline{Underlined} = second-highest}}
\end{tabular}
\end{table*}

\subsection{Prompt-wise Toxicity Prediction Performance}
We evaluated the effectiveness of various prompting strategies for toxicity prediction using GPT-4o, as summarized in Table I. Among the settings that utilize only chemical structure as input, Zeroshot prompts yielded the lowest average F1-scores (0.370 for SMILES and 0.368 for IUPAC), indicating limited predictive ability when no examples or reasoning are provided. Incorporating few-shot examples improved performance to 0.434 for SMILES and 0.402 for IUPAC, highlighting the benefit of in-context learning. CoT prompting, which guides the model to generate step-by-step reasoning, resulted in modest gains for IUPAC (0.417), but did not outperform the Fewshot setting in the SMILES case (0.409).\\
\indent Notably, the choice between SMILES and IUPAC representations had only a marginal impact under chemical-only settings, with performance differences remaining within ±0.03 across all prompt types. This suggests that structural format alone does not significantly influence prediction quality in the absence of biological context.\\
\indent In contrast, when biological process information was used independently (BioProcess-CoT), performance improved across most toxicity types compared to structure-only prompts. This result highlights the importance of biological information, such as pathway and GO term annotations, in predicting organ-specific toxicity. In particular, hematological and liver toxicity showed the greatest improvements when using biological information alone, indicating that biological mechanisms may be more informative than structural patterns for these toxicity types.\\
\indent Building on this, CoTox further enhanced performance by combining both chemical structure and biological context within a unified reasoning framework. CoTox (SMILES) achieved an average F1-score of 0.659, and CoTox slightly outperformed it at 0.663. This represents a performance gain of over 0.25 compared to structure-CoT-only prompts, and a clear improvement over traditional baselines such as XGBoost (0.576) and Chemprop (0.619).\\
\indent Across the six toxicity types, hematological and liver toxicity showed the highest predictive scores (up to 0.817 and 0.774, respectively), while pulmonary and renal toxicity remained the most challenging, with F1-scores consistently below 0.57 across methods. These results underscore the value of combining chemical and biological context, and demonstrate how well-crafted prompts can enable LLMs to outperform conventional approaches.

\vspace{-1.2mm}
\subsection{Model-wise Toxicity Prediction Performance}
To assess the effect of language model architecture on toxicity prediction, we tested a diverse set of general, expert, and reasoning LLMs using the same CoTox prompting strategy. All models were provided with IUPAC representations as structural input, while SMILES-based results were included only as average scores for comparison. The performance gap between IUPAC and SMILES inputs is also reported to highlight the effect of structural format within the CoTox framework. All experiments were
done with Nvidia A100 80GB, adopting a greedy decoding with
temperature 0.0. Table II summarizes their performance across six toxicity types.

Gemini-2.5-Pro achieved the highest overall performance, with an average F1-score of 0.700, outperforming all other models. It demonstrated consistently strong results across toxicity types, including top scores in cardio (0.746), infertility (0.630), and renal toxicity (0.606). GPT-4o and Llama3.1-8B also performed well, achieving average F1-scores of 0.663 and 0.685, respectively, with particularly strong results in hematological and liver toxicity.

In contrast, Llama3.1-70B, despite achieving the highest hematological toxicity score (0.835), exhibited unstable performance, scoring lowest on infertility (0.292) and pulmonary toxicity (0.391), resulting in an overall average of 0.615. Although the expert model, TxGemma, was fine-tuned on a large and diverse set of biomedical tasks from the Therapeutic Data Commons (TDC), it showed the weakest performance overall (0.387), suggesting that the model, having been primarily trained to infer molecular properties from chemical structures, may have struggled to interpret biological information such as pathway and GO term annotations. This limitation likely hindered its ability to integrate structural and biological context effectively in the CoTox prompting framework.

Reasoning LLMs such as o3, Qwen3-32B, and DeepSeek-R1 (distilled from Llama3.1-70B) demonstrated notable improvements compared to their SMILES-only baselines, with gap improvements of +15.6\%, +15.02\%, and +10.24\%, respectively. These results suggest that the human-readable nature of IUPAC may better align with the reasoning capabilities of instruction-tuned models than tokenized formats like SMILES. This alignment likely enables models to more effectively parse and relate structural information to biological context within multi-step inference frameworks such as CoTox. In contrast, Gemini-2.5-Pro, while achieving the highest overall performance, exhibited only a marginal difference between IUPAC and SMILES inputs. This suggests that the model may already possess a strong ability to infer chemical structure from SMILES, effectively bridging the gap to the explicit representations provided by IUPAC. Such capacity indicates a deeper structural understanding embedded within the model, reducing its dependence on more interpretable formats.

Overall, these results imply that LLM reasoning ability plays a critical role in CoTox performance, with models capable of integrating biological context and chemical structure showing superior predictive power across diverse toxicity endpoints.

\subsection{Impact of Structural Representations: SMILES vs. IUPAC}
As shown in Table II, all models except Llama3.1-70B exhibited improved performance when provided with IUPAC representations instead of SMILES. The performance gap was most pronounced in reasoning-focused models, where the IUPAC input led to substantial improvements. This trend suggests that structurally descriptive and linguistically aligned formats like IUPAC better support the multi-step reasoning capabilities of instruction-tuned LLMs.

The SMILES format, while compact and machine-readable, often lacks the semantic richness and structural clarity found in IUPAC nomenclature. In contrast, IUPAC names encapsulate chemically meaningful features such as ring systems, functional groups, and positional information in natural language terms. This allows instruction-tuned LLMs to extract relevant structural clues and link them to biological or toxicological outcomes more effectively.
\begin{figure}[!ht]
    \centering
    \includegraphics[width=\linewidth]{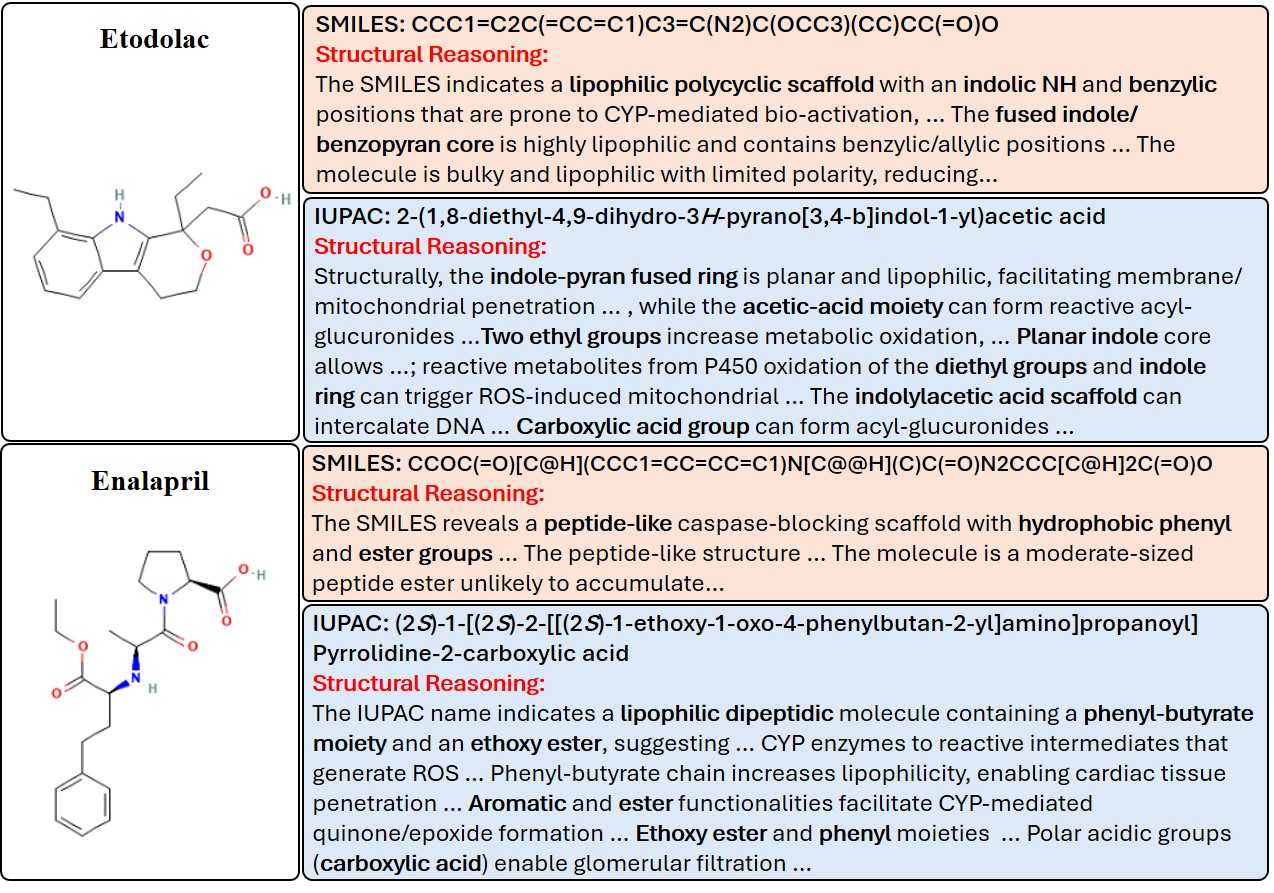}
    \caption{Comparison of Structural reasoning using SMILES vs IUPAC input.}
    \label{fig_2}
    \vspace{-15pt}
\end{figure}
A qualitative comparison in Fig. \ref{fig_2}., based on the structural reasoning segments extracted from the full CoTox prompt outputs of o3, illustrates this point. For the same compound, Etodolac, the SMILES representation vaguely describes atomic connectivity, whereas the IUPAC representation explicitly conveys the presence of an ‘indole-pyran fused ring’ and a ‘carboxylic acid group’, enabling more detailed reasoning about lipophilicity, bioactivation potential, and organ-level toxicity. Similarly, for Enalapril, IUPAC enables the model to identify a ‘phenylbutyrate’ and ‘dipeptidic’ structure, leading to more accurate mapping of structural motifs to potential toxicity pathways.

These findings underscore that while both SMILES and IUPAC encode chemical structure, the interpretability and linguistic alignment of IUPAC make it a more effective input modality for CoTox-style chain-of-thought toxicity prediction.

\subsection{Case Study: Reasoning Outcome Analysis}
\begin{table}[t]
\footnotesize
\centering
\caption{Case Study of Propranolol}
\label{case_1}
\renewcommand{\arraystretch}{1.0}
\begin{tabular}{p{0.45\textwidth}}
\toprule
\textbf{\textit{Case Study: Propranolol by Gemini-2.5-Pro with CoTox}}
\\\\
\textbf{Toxicity Prediction}\\
\vspace{2pt}
\textbf{``Cardiotoxicity":}\\ 
\{Reasoning: \\

\textit{Pathway}: The provided pathways, such as \hlpathway{`Intrinsic Pathway for Apoptosis'} and \hl{`Activation of caspases'}, describe general cell death mechanisms...\\
\textit{GO Term}: General GO terms like \hlgo{`positive regulation of oxidative stress-induced cell death'} can apply to cardiac tissue...\\
\textit{IUPAC Support}: The \hlstruct{`propanolamine side chain'}, is responsible for binding to beta-adrenergic receptors in the heart. ...\\
\textit{Overall Mechanism}: As a beta-blocker, the compound reduces heart rate and contractility ... trigger apoptotic pathways in cardiomyocytes, leading to cell death ...\},\\
\{Prediction: \textcolor{red}{``Toxic"}\}, \{Answer: \textcolor{red}{``Toxic"}\}\\\\

\textbf{``Hematological Toxicity":}\\ 

\{Reasoning: ... ... \},\\
\{Prediction: \textcolor{red}{``Toxic"}\}, \{Answer: \textcolor{red}{``Toxic"}\}\\\\

\textbf{``Infertility":}\\ 
\{Reasoning: ... ... \},\\
\{Prediction: \textcolor{blue}{``Non-toxic"}\}, \{Answer: \textcolor{blue}{``Non-toxic"}\}\\\\

\textbf{``Liver Toxicity":}\\ 
\{Reasoning: \\
\textit{Pathway}: The \hlpathway{`CYP2E1 reactions'} and \hl{`Xenobiotics'} pathways are direct indicators of liver toxicity ... which then trigger the \hl{`Intrinsic Pathway for Apoptosis'} and \hl{`Cytochrome c-mediated apoptotic response'} in hepatocytes\\

\textit{GO Term}: Terms such as \hlgo{`xenobiotic catabolic process'} and \hl{`toxin metabolic process'} ... that leads to oxidative stress and cell death.\\

\textit{IUPAC Support}: The \hlstruct{`naphthalene ring'} is a polycyclic aromatic hydrocarbon (PAH) moiety ... structures by P450 enzymes like CYP2E1 can form reactive epoxide intermediates.\\

\textit{Overall Mechanism}: These combined effects overwhelm antioxidant defenses, initiating apoptosis and culminating in drug-induced liver injury.\},\\
\{Prediction: \textcolor{red}{``Toxic"}\}, \{Answer: \textcolor{red}{``Toxic"}\}\\\\

\textbf{``Pulmonary Toxicity":}\\ 
\{Reasoning: ... ... \},\\
\{Prediction: \textcolor{red}{``Toxic"}\}, \{Answer: \textcolor{red}{``Toxic"}\}\\\\

\textbf{``Renal Toxicity":}\\ 
\{Reasoning: \\
\textit{Pathway}: No pathways specific to renal function, ...\\

\textit{GO Term}: There are no GO terms related to the kidney ...\\

\textit{IUPAC Support}: The structure does not possess ... structural alerts for nephrotoxicity (e.g., aminoglycosides, platinum-containing compounds).\\

\textit{Overall Mechanism}: The provided data from pathways and GO terms offers no evidence to suggest a mechanism for renal toxicity. ...\},\\
\{Prediction: \textcolor{blue}{``Non-toxic"}\}, \{Answer: \textcolor{blue}{``Non-toxic"}\}\\\\

\bottomrule
\end{tabular}
\vspace{7pt}
\end{table}
To assess the reliability of LLM-based toxicity reasoning, we analyzed the predicted toxicities and supporting explanations generated by Gemini-2.5-Pro with CoTox for the drug Propranolol, as shown in Table \ref{case_1}. Each toxicity type is evaluated in terms of pathway and GO term alignment, as well as structural interpretation based on IUPAC.\\
For cardiotoxicity, the model’s reference to `intrinsic apoptotic pathways' and `oxidative stress' is valid, as prior work shows that propranolol and similar beta-blockers can activate mitochondrial apoptosis and increase ROS levels in cardiac cells\cite{zhao2020propranolol}. The structural explanation, involving propranolol’s beta-adrenergic blocking action, also fits with known mechanisms of reduced contractility and pro-apoptotic signaling in the heart\cite{frishman2013beta}. Regarding liver toxicity, the model’s emphasis on `CYP2E1-mediated metabolism' and `xenobiotic' processing correctly reflects the mechanism by which drugs containing a `naphthalene ring', a bicyclic fused aromatic hydrocarbon, undergo enzymatic oxidation. Propranolol’s chemical structure includes a naphthalene-like moiety which can be metabolized by CYP450 enzymes (including CYP2E1), forming reactive metabolites such as epoxides and quinones. These metabolites are well-documented to induce oxidative stress, lipid peroxidation, and hepatocyte damage, contributing to liver toxicity\cite{guengerich2008cytochrome, makar2019naphthalene}. GO terms such as `xenobiotic catabolic process' accurately represent the main biochemical events in drug-induced liver injury\cite{sturgill1997xenobiotic}. For renal toxicity, the model’s reasoning is supported by the lack of strong evidence linking propranolol to kidney damage, as it is neither known to affect major renal pathways nor does its structure resemble classic nephrotoxins\cite{thompson1972pharmacodynamics}.

\vspace{-1mm}

\subsection{Case Study: Organ-Specific Cell Line Guided Reasoning}

In this case study, we demonstrate a method for predicting toxicity by inputting pathway and GO terms derived from gene expression changes induced by drug treatment in organ-specific cell lines into the CoTox prompt. The CoTox framework described in earlier sections relies on pre-existing biological context retrieved from public databases. However, such information is often unavailable for newly developed compounds. To overcome this limitation, gene expression changes can be obtained by experimentally treating relevant cell lines with the compound or by using predictive models\cite{ qi2024predicting}. These changes can then be analyzed through GSEA(Gene Set Enrichment Analysis)\cite{subramanian2005gene} to extract biologically meaningful pathways and GO terms.

Using this approach, CoTox enables organ-specific toxicity prediction and mechanistic reasoning, even in the absence of prior biological knowledge. This highlights the practical value of the framework in early-stage drug development.
In Table \ref{case_2}, we investigate the toxicity of the drug Entecavir across three organs: liver, lung, and kidney. The ground-truth toxicity labels were determined based on information from the FDA Drug Label, following the same methodology used in the UniTox dataset.

To obtain organ-specific biological context, we utilized L1000 Level 5 gene expression profiles\cite{subramanian2017next} induced by Entecavir. For liver, we used the HEPG2 cell line; for lung, A549; and for kidney, HA1E. GSEA was performed on each expression profile to extract statistically significant pathways and GO terms, using a filtering threshold of FDR q-value $< 0.25$ and p-value $< 0.01$, which are commonly accepted cutoffs for statistical significance in high-throughput analyses.
Subsequently, as in the previous CoTox setup, we applied GPT-4o to filter out toxicity-relevant biological terms from the enriched results. These curated pathway and GO terms were input into the CoTox prompt, and toxicity predictions along with mechanistic reasoning were generated using the Gemini-2.5-Pro.
\begin{table}[t]
\footnotesize
\centering
\caption{Case Study of Entecavir}
\label{case_2}
\renewcommand{\arraystretch}{1.0}
\begin{tabular}{p{0.45\textwidth}}
\toprule
\textbf{\textit{Case Study: Entecavir by Gemini-2.5-Pro with CoTox}}
\\\\
\textbf{Toxicity Prediction}\\
\vspace{2pt}

\textbf{``Liver Toxicity":}\\ 
\{Reasoning: \\
\textit{Pathway}: No pathway provided.\\

\textit{GO Term}: The provided GO terms (\hlgo{`Transmembrane Receptor Protein Tyrosine Phosphatase Activity', `Protein Serine/Threonine/Tyrosine Kinase Activity'}) indicate interference with cellular phosphorylation and dephosphorylation. Dysregulation of these fundamental signaling cascades is a known mechanism of drug-induced liver injury (DILI), as it can disrupt cell growth, proliferation, and survival, leading to hepatocyte apoptosis and inflammation.\\

\textit{IUPAC Support}: The structure is a \hlstruct{`guanosine analog'}. For its therapeutic action, it must be phosphorylated by host cellular kinases. ... lead to off-target effects, disrupting the delicate balance of kinase and phosphatase signaling networks, ... contributing to hepatotoxicity.\\

\textit{Overall Mechanism}: The compound, a guanosine analog, interacts with cellular kinases for its bioactivation. ... leads to cellular stress and apoptosis, culminating in drug-induced liver injury.\},\\
\{Prediction: \textcolor{red}{``Toxic"}\}, \{Answer: \textcolor{red}{``Toxic"}\}\\\\

\textbf{``Pulmonary Toxicity":}\\ 
\{Reasoning: \\
\textit{Pathway}: No pathway provided.\\
\textit{GO Term}: No pathway provided.\\
\textit{IUPAC Support}: The chemical structure, a \hlstruct{`purine nucleoside analog'}, does not contain functional groups or structural motifs commonly associated with direct pulmonary toxicity, such as reactive electrophiles or moieties known to generate significant oxidative stress specifically in lung tissue.\\
\textit{Overall Mechanism}: In the absence of any implicated pathways or biological processes from the provided data, there is no evidence to construct a plausible mechanism for pulmonary toxicity.\},\\
\{Prediction: \textcolor{blue}{``Non-toxic"}\}, \{Answer: \textcolor{blue}{``Non-toxic"}\}\\\\

\textbf{``Renal Toxicity":}\\ 
\{Reasoning: \\
\textit{Pathway}: The involvement of \hlpathway{`p53-Dependent G1 S DNA Damage Checkpoint' and `p53-Dependent G1 DNA Damage Response'} pathways is a strong ... Activation of these pathways signifies that the compound is causing DNA damage in renal cells, leading to cell cycle arrest and potentially apoptosis to eliminate damaged cells.\\

\textit{GO Term}: No GO terms provided\\

\textit{IUPAC Support}: The molecule is a \hlstruct{`nucleoside analog'} designed to interfere with DNA synthesis. ... accumulation can cause off-target effects on host cell DNA replication, which is recognized as DNA damage, thereby triggering the p53-mediated response ... \\

\textit{Overall Mechanism}: As a nucleoside analog, the compound can accumulate in renal proximal tubules. ... DNA damage ... p53-dependent DNA damage response. This response triggers cell cycle arrest and apoptosis in renal tubular cells, ..  and overall nephrotoxicity.\},\\
\{Prediction: \textcolor{red}{``Toxic"}\}, \{Answer: \textcolor{blue}{``Non-toxic"}\}\\

\bottomrule
\end{tabular}
\vspace{7pt}
\end{table}
Among the three target organs, CoTox correctly predicted liver and pulmonary toxicity outcomes, aligning with the known safety profiles of Entecavir. However, for renal toxicity, CoTox predicted a toxic outcome despite the ground truth label indicating non-toxicity. Interestingly, recent clinical evidence suggests that Entecavir may indeed pose a risk of renal function decline. A clinical study found that patients with chronic hepatitis B who were treated with Entecavir had a higher chance of kidney function decline compared to those treated with tenofovir alafenamide (adjusted hazard ratio 4.05; p $< .001$)\cite{jung2022higher}. These findings indicate that CoTox may capture latent toxicity signals not yet fully represented in regulatory documents, underscoring its potential utility in early toxicity risk assessment.

\section{Conclusion}
In this study, we introduced CoTox, a toxicity reasoning framework that leverages large language models to integrate chemical structures, pathway data, and GO terms for multi-organ toxicity prediction. CoTox outperformed traditional ML/DL models and prior prompt-based methods, offering improved performance and interpretable reasoning grounded in biological mechanisms.

Our analysis showed that IUPAC names better align with LLMs’ language understanding than SMILES, suggesting practical benefits for chemical input formatting. Through case studies, we confirmed that CoTox’s explanations matched known toxicological pathways and literature evidence. Moreover, by incorporating gene expression profiles from organ-specific cell lines, CoTox demonstrated the ability to infer toxicity even for unannotated compounds.

These findings position CoTox as an interpretable and practical tool for early-stage drug development. Future work may further enhance its robustness by integrating pharmacological data, dose-response effects, and multimodal biological inputs.

\section{Acknowledgments}
This work was supported by National Supercomputing Center with supercomputing resources including technical support (KSC-2024-CRE-0391).

\bibliographystyle{IEEEtran}

\begin{thebibliography}{10}
\providecommand{\url}[1]{#1}
\csname url@samestyle\endcsname
\providecommand{\newblock}{\relax}
\providecommand{\bibinfo}[2]{#2}
\providecommand{\BIBentrySTDinterwordspacing}{\spaceskip=0pt\relax}
\providecommand{\BIBentryALTinterwordstretchfactor}{4}
\providecommand{\BIBentryALTinterwordspacing}{\spaceskip=\fontdimen2\font plus
\BIBentryALTinterwordstretchfactor\fontdimen3\font minus \fontdimen4\font\relax}
\providecommand{\BIBforeignlanguage}[2]{{%
\expandafter\ifx\csname l@#1\endcsname\relax
\typeout{** WARNING: IEEEtran.bst: No hyphenation pattern has been}%
\typeout{** loaded for the language `#1'. Using the pattern for}%
\typeout{** the default language instead.}%
\else
\language=\csname l@#1\endcsname
\fi
#2}}
\providecommand{\BIBdecl}{\relax}
\BIBdecl

\bibitem{sun202290}
D.~Sun, W.~Gao, H.~Hu, and S.~Zhou, ``Why 90\% of clinical drug development fails and how to improve it?'' \emph{Acta Pharmaceutica Sinica B}, vol.~12, no.~7, pp. 3049--3062, 2022.

\bibitem{babai2021safety}
S.~Babai, L.~Auclert, and H.~Le-Lou{\"e}t, ``Safety data and withdrawal of hepatotoxic drugs,'' \emph{Therapies}, vol.~76, no.~6, pp. 715--723, 2021.

\bibitem{tran2023artificial}
T.~T.~V. Tran, A.~Surya~Wibowo, H.~Tayara, and K.~T. Chong, ``Artificial intelligence in drug toxicity prediction: recent advances, challenges, and future perspectives,'' \emph{Journal of chemical information and modeling}, vol.~63, no.~9, pp. 2628--2643, 2023.

\bibitem{patil2025advancing}
A.~Patil and A.~Jadon, ``Advancing reasoning in large language models: Promising methods and approaches,'' \emph{arXiv preprint arXiv:2502.03671}, 2025.

\bibitem{zheng2024large}
Y.~Zheng, H.~Y. Koh, M.~Yang, L.~Li, L.~T. May, G.~I. Webb, S.~Pan, and G.~Church, ``Large language models in drug discovery and development: From disease mechanisms to clinical trials,'' \emph{arXiv preprint arXiv:2409.04481}, 2024.

\bibitem{yang2025large}
H.~Yang, J.~Xiu, W.~Yan, K.~Liu, H.~Cui, Z.~Wang, Q.~He, Y.~Gao, and W.~Han, ``Large language models as tools for molecular toxicity prediction: Ai insights into cardiotoxicity,'' \emph{Journal of Chemical Information and Modeling}, vol.~65, no.~5, pp. 2268--2282, 2025.

\bibitem{chen2025application}
Y.-Q. Chen, T.~Yu, Z.-Q. Song, C.-Y. Wang, J.-T. Luo, Y.~Xiao, H.~Qiu, Q.-Q. Wang, and H.-M. Jin, ``Application of large language models in drug-induced osteotoxicity prediction,'' \emph{Journal of Chemical Information and Modeling}, vol.~65, no.~7, pp. 3370--3379, 2025.

\bibitem{jang2025improving}
Y.~Jang, J.~Kim, and S.~Ahn, ``Improving chemical understanding of llms via smiles parsing,'' \emph{arXiv preprint arXiv:2505.16340}, 2025.

\bibitem{lin2019off}
A.~Lin, C.~J. Giuliano, A.~Palladino, K.~M. John, C.~Abramowicz, M.~L. Yuan, E.~L. Sausville, D.~A. Lukow, L.~Liu, A.~R. Chait \emph{et~al.}, ``Off-target toxicity is a common mechanism of action of cancer drugs undergoing clinical trials,'' \emph{Science translational medicine}, vol.~11, no. 509, p. eaaw8412, 2019.

\bibitem{chen2016analysis}
L.~Chen, Y.-H. Zhang, Q.~Zou, C.~Chu, and Z.~Ji, ``Analysis of the chemical toxicity effects using the enrichment of gene ontology terms and kegg pathways,'' \emph{Biochimica et Biophysica Acta (BBA)-General Subjects}, vol. 1860, no.~11, pp. 2619--2626, 2016.

\bibitem{wei2022chain}
J.~Wei, X.~Wang, D.~Schuurmans, M.~Bosma, F.~Xia, E.~Chi, Q.~V. Le, D.~Zhou \emph{et~al.}, ``Chain-of-thought prompting elicits reasoning in large language models,'' \emph{Advances in neural information processing systems}, vol.~35, pp. 24\,824--24\,837, 2022.

\bibitem{ramos2025review}
M.~C. Ramos, C.~J. Collison, and A.~D. White, ``A review of large language models and autonomous agents in chemistry,'' \emph{Chemical science}, 2025.

\bibitem{chatgpt}
OpenAI, ``Introducing chatgpt,'' 2022.

\bibitem{ouyang2023structured}
S.~Ouyang, Z.~Zhang, B.~Yan, X.~Liu, Y.~Choi, J.~Han, and L.~Qin, ``Structured chemistry reasoning with large language models,'' \emph{arXiv preprint arXiv:2311.09656}, 2023.

\bibitem{xian2025molrag}
Z.~Xian, J.~Gu, L.~Li, and S.~Liang, ``Molrag: unlocking the power of large language models for molecular property prediction,'' in \emph{Proceedings of the 63rd Annual Meeting of the Association for Computational Linguistics (Volume 1: Long Papers)}, 2025, pp. 15\,513--15\,531.

\bibitem{fallahpour2025bioreason}
A.~Fallahpour, A.~Magnuson, P.~Gupta, S.~Ma, J.~Naimer, A.~Shah, H.~Duan, O.~Ibrahim, H.~Goodarzi, C.~J. Maddison \emph{et~al.}, ``Bioreason: Incentivizing multimodal biological reasoning within a dna-llm model,'' \emph{arXiv preprint arXiv:2505.23579}, 2025.

\bibitem{davis2025comparative}
A.~P. Davis, T.~C. Wiegers, D.~Sciaky, F.~Barkalow, M.~Strong, B.~Wyatt, J.~Wiegers, R.~McMorran, S.~Abrar, and C.~J. Mattingly, ``Comparative toxicogenomics database’s 20th anniversary: update 2025,'' \emph{Nucleic acids research}, vol.~53, no.~D1, pp. D1328--D1334, 2025.

\bibitem{kim2018update}
S.~Kim, P.~A. Thiessen, T.~Cheng, B.~Yu, and E.~E. Bolton, ``An update on pug-rest: Restful interface for programmatic access to pubchem,'' \emph{Nucleic Acids Research}, vol.~46, no.~W1, pp. W563--W570, 2018.

\bibitem{silberg2024unitox}
J.~Silberg, K.~Swanson, E.~Simon, A.~Zhang, Z.~Ghazizadeh, S.~Ogden, H.~Hamadeh, and J.~Y. Zou, ``Unitox: leveraging llms to curate a unified dataset of drug-induced toxicity from fda labels,'' \emph{Advances in Neural Information Processing Systems}, vol.~37, pp. 12\,078--12\,093, 2024.

\bibitem{chen2016xgboost}
T.~Chen and C.~Guestrin, ``Xgboost: A scalable tree boosting system,'' in \emph{Proceedings of the 22nd acm sigkdd international conference on knowledge discovery and data mining}, 2016, pp. 785--794.

\bibitem{heid2023chemprop}
E.~Heid, K.~P. Greenman, Y.~Chung, S.-C. Li, D.~E. Graff, F.~H. Vermeire, H.~Wu, W.~H. Green, and C.~J. McGill, ``Chemprop: a machine learning package for chemical property prediction,'' \emph{Journal of Chemical Information and Modeling}, vol.~64, no.~1, pp. 9--17, 2023.

\bibitem{gpt4o}
OpenAI, ``Hello gpt-4o,'' 2024.

\bibitem{Llama3.1}
Meta, ``Introducing llama 3.1: Our most capable models to date,'' 2024.

\bibitem{wang2025txgemma}
E.~Wang, S.~Schmidgall, P.~F. Jaeger, F.~Zhang, R.~Pilgrim, Y.~Matias, J.~Barral, D.~Fleet, and S.~Azizi, ``Txgemma: Efficient and agentic llms for therapeutics,'' \emph{arXiv preprint arXiv:2504.06196}, 2025.

\bibitem{o3}
OpenAI, ``Openai o3 and o4-mini system card,'' 2025.

\bibitem{deepseek-r1}
D.~Guo, D.~Yang, H.~Zhang, J.~Song, R.~Zhang, R.~Xu, Q.~Zhu, S.~Ma, P.~Wang, X.~Bi \emph{et~al.}, ``Deepseek-r1: Incentivizing reasoning capability in llms via reinforcement learning,'' \emph{arXiv preprint arXiv:2501.12948}, 2025.

\bibitem{qwen3}
A.~Yang, A.~Li, B.~Yang, B.~Zhang, B.~Hui, B.~Zheng, B.~Yu, C.~Gao, C.~Huang, C.~Lv \emph{et~al.}, ``Qwen3 technical report,'' \emph{arXiv preprint arXiv:2505.09388}, 2025.

\bibitem{gemini2.5}
G.~Comanici, E.~Bieber, M.~Schaekermann, I.~Pasupat, N.~Sachdeva, I.~Dhillon, M.~Blistein, O.~Ram, D.~Zhang, E.~Rosen \emph{et~al.}, ``Gemini 2.5: Pushing the frontier with advanced reasoning, multimodality, long context, and next generation agentic capabilities,'' \emph{arXiv preprint arXiv:2507.06261}, 2025.

\bibitem{zhao2020propranolol}
S.~Zhao, S.~Fan, Y.~Shi, H.~Ren, H.~Hong, X.~Gao, M.~Zhang, Q.~Qin, and H.~Li, ``Propranolol induced apoptosis and autophagy via the ros/jnk signaling pathway in human ovarian cancer,'' \emph{Journal of Cancer}, vol.~11, no.~20, p. 5900, 2020.

\bibitem{frishman2013beta}
W.~H. Frishman, ``$\beta$-adrenergic blockade in cardiovascular disease,'' \emph{Journal of cardiovascular pharmacology and therapeutics}, vol.~18, no.~4, pp. 310--319, 2013.

\bibitem{guengerich2008cytochrome}
F.~P. Guengerich, ``Cytochrome p450 and chemical toxicology,'' \emph{Chemical research in toxicology}, vol.~21, no.~1, pp. 70--83, 2008.

\bibitem{makar2019naphthalene}
S.~Makar, T.~Saha, and S.~K. Singh, ``Naphthalene, a versatile platform in medicinal chemistry: Sky-high perspective,'' \emph{European journal of medicinal chemistry}, vol. 161, pp. 252--276, 2019.

\bibitem{sturgill1997xenobiotic}
M.~G. Sturgill and G.~H. Lambert, ``Xenobiotic-induced hepatotoxicity: mechanisms of liver injury and methods of monitoring hepatic function,'' \emph{Clinical chemistry}, vol.~43, no.~8, pp. 1512--1526, 1997.

\bibitem{thompson1972pharmacodynamics}
F.~Thompson, A.~Joekes, and D.~Foulkes, ``Pharmacodynamics of propranolol in renal failure,'' \emph{Br Med J}, vol.~2, no. 5811, pp. 434--436, 1972.

\bibitem{qi2024predicting}
X.~Qi, L.~Zhao, C.~Tian, Y.~Li, Z.-L. Chen, P.~Huo, R.~Chen, X.~Liu, B.~Wan, S.~Yang \emph{et~al.}, ``Predicting transcriptional responses to novel chemical perturbations using deep generative model for drug discovery,'' \emph{Nature Communications}, vol.~15, no.~1, p. 9256, 2024.

\bibitem{subramanian2005gene}
A.~Subramanian, P.~Tamayo, V.~K. Mootha, S.~Mukherjee, B.~L. Ebert, M.~A. Gillette, A.~Paulovich, S.~L. Pomeroy, T.~R. Golub, E.~S. Lander \emph{et~al.}, ``Gene set enrichment analysis: a knowledge-based approach for interpreting genome-wide expression profiles,'' \emph{Proceedings of the National Academy of Sciences}, vol. 102, no.~43, pp. 15\,545--15\,550, 2005.

\bibitem{subramanian2017next}
A.~Subramanian, R.~Narayan, S.~M. Corsello, D.~D. Peck, T.~E. Natoli, X.~Lu, J.~Gould, J.~F. Davis, A.~A. Tubelli, J.~K. Asiedu \emph{et~al.}, ``A next generation connectivity map: L1000 platform and the first 1,000,000 profiles,'' \emph{Cell}, vol. 171, no.~6, pp. 1437--1452, 2017.

\bibitem{jung2022higher}
C.-Y. Jung, H.~W. Kim, S.~H. Ahn, S.~U. Kim, and B.~S. Kim, ``Higher risk of kidney function decline with entecavir than tenofovir alafenamide in patients with chronic hepatitis b,'' \emph{Liver International}, vol.~42, no.~5, pp. 1017--1026, 2022.

\end{thebibliography}

\end{document}